%% file: _main.tex
\input{_constants}

\pdfoutput=1
\documentclass[10pt,twocolumn,letterpaper]{article}
\input{cvpr_header}

\unless\ifarxiv \myexternaldocument{_supplementary} \fi

\begin{document}
\title{\paperTitle}
\author{\authorBlock}
\maketitle

\input{00_abstract}
\input{01_intro}
\input{02_related}

\input{03_method}

\input{04_experi}

\input{10_conclusion}

\input{13_references}
{\small
}

\ifarxiv \clearpage \appendix \input{12_appendix} \fi

\end{document}

%% file: _constants.tex
\def\paperID{0000}
\def\confName{ICCV}
\def\confYear{2025}

\def\paperTitle{Robust Object Detection for Autonomous Driving via Curriculum-Guided Group Relative Policy Optimization}

\def\authorBlock{
    Xu Jia \\
    Shandong Normal University\\
    {\tt\small 202311900124@sdnu.edu.cn}
}

\newif\ifreview 
\newif\ifarxiv 
\newif\ifcamera 
\newif\ifrebuttal 

%% file: cvpr_header.tex
\ifreview \usepackage[review]{cvpr} \fi
\ifarxiv \usepackage[pagenumbers]{cvpr} \fi
\ifrebuttal \usepackage[rebuttal]{cvpr} \fi
\ifcamera \usepackage{cvpr} \fi

\input{_macros}  

\usepackage{xr-hyper}

\makeatletter
\newcommand*{\addFileDependency}[1]{
  \typeout{(#1)}
  \@addtofilelist{#1}
  \IfFileExists{#1}{}{\typeout{No file #1.}}
}

\makeatother
\newcommand*{\myexternaldocument}[1]{
    \externaldocument{#1}
    \addFileDependency{#1.tex}
    \addFileDependency{#1.aux}
}

\definecolor{cvprblue}{rgb}{0.21,0.49,0.74}
\usepackage[pagebackref,breaklinks,colorlinks,allcolors=cvprblue]{hyperref}
\usepackage[capitalize]{cleveref}
\crefname{section}{Sec.}{Secs.}
\crefname{table}{Table}{Tables}
\crefname{figure}{Fig.}{Figs.}

\ifarxiv \crefname{appendix}{App.}{Apps.}
\else \crefname{appendix}{Suppl.}{Suppls.} \fi

%% file: _macros.tex
\usepackage{graphicx}	
\usepackage{amsmath}	
\usepackage{amssymb}	
\usepackage{booktabs}
\usepackage{times}
\usepackage{microtype}
\usepackage{epsfig}
\usepackage{caption}
\usepackage{float}
\usepackage{placeins}
\usepackage{color, colortbl}
\usepackage{stfloats}
\usepackage{enumitem}
\usepackage{tabularx}
\usepackage{xstring}
\usepackage{multirow}
\usepackage{xspace}
\usepackage{url}
\usepackage{subcaption}
\usepackage{xcolor}
\usepackage[hang,flushmargin]{footmisc}

\ifcamera \usepackage[accsupp]{axessibility} \fi

% ...

\ifarxiv  \fi

\newcommand{\R}[1]{{%
    \textbf{%
        \ifstrequal{#1}{1}{\textcolor{red}{R#1}}{%
        \ifstrequal{#1}{2}{\textcolor{blue}{R#1}}{%
        \ifstrequal{#1}{3}{\textcolor{magenta}{R#1}}{%
        \ifstrequal{#1}{4}{\textcolor{teal}{R#1}}{%
                           \textcolor{cyan}{R#1}%
        }}}}%
    }%
}}

%% file: 00_abstract.tex
\begin{abstract}
Multimodal Large Language Models (MLLMs) excel in vision-language reasoning but often struggle with structured perception tasks requiring precise localization and robustness. We propose a reinforcement learning framework that augments Group Relative Policy Optimization (GRPO) with curriculum-based data scheduling and difficulty-aware filtering. This approach stabilizes optimization under sparse, noisy rewards and enables progressive adaptation to complex samples. Evaluations on autonomous driving benchmarks demonstrate substantial improvements in detection accuracy and robustness. Ablation studies confirm the importance of reward design, KL regularization, and curriculum pacing for convergence stability and generalization. Our findings highlight reinforcement-driven optimization with structured data curricula as a scalable path toward robust and interpretable multimodal detection.
\begin{center}
\end{center}

\end{abstract}

%% file: 01_intro.tex
\section{Introduction}
\label{sec:intro}
The emergence of MLLMs has significantly advanced the frontier of vision-language understanding, enabling strong performance on tasks like visual reasoning and dialogue [1, 2]. Through large-scale pretraining and unified modeling[3, 4] , these models begin to approximate human-level multimodal cognition. Nevertheless, their ability to perform structured perception—where
\begin{figure}[t]
\centering
\includegraphics[width=\linewidth,keepaspectratio]{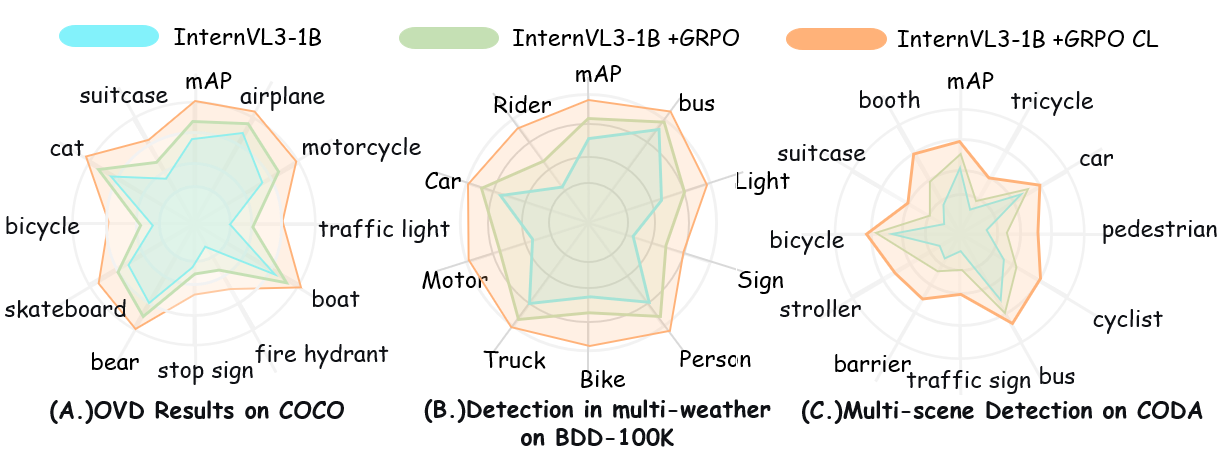}
\end{figure}

\begin{figure}[t]
\centering
\includegraphics[width=\linewidth,keepaspectratio]{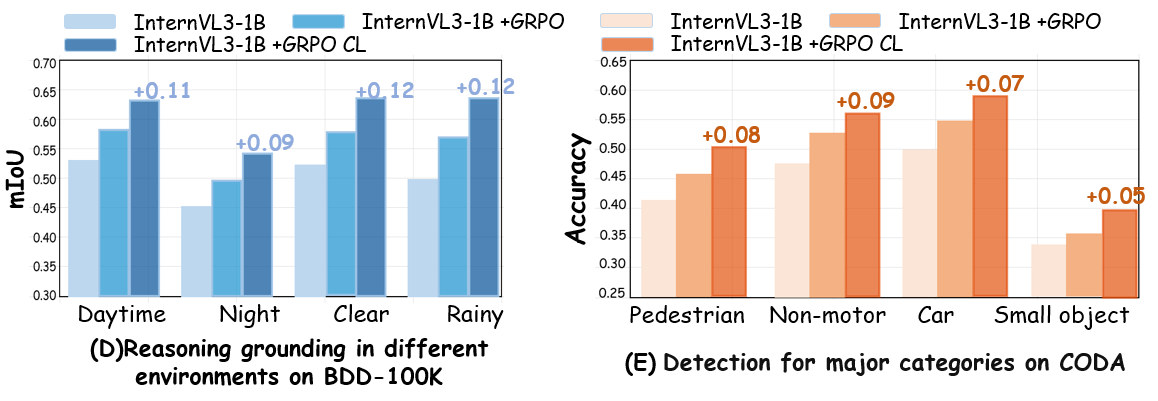}
\end{figure}

 accurate spatial localization and semantic grounding are essential—remains limited[5, 6]. In particular, MLLMs exhibit vulnerability to data imbalances and labeling noise, and often lack the granularity required for tasks such as region-level comprehension and detection. This gap poses a critical barrier to their adoption in high-stakes a plications, including autonomous driving.\\
To address these limitations, recent efforts have explored integrating reinforcement learning (RL) paradigms into the optimization of MLLMs. Among them, GRPO has emerged as a promising approach, improving the stability and efficiency of policy updates through relative comparisons within reward-ranked groups. However, GRPO still faces practical challenges: reward signals in multimodal perception tasks are often 
sparse and 
\begin{figure*}[t]
\centering
\includegraphics[width=\textwidth,keepaspectratio]{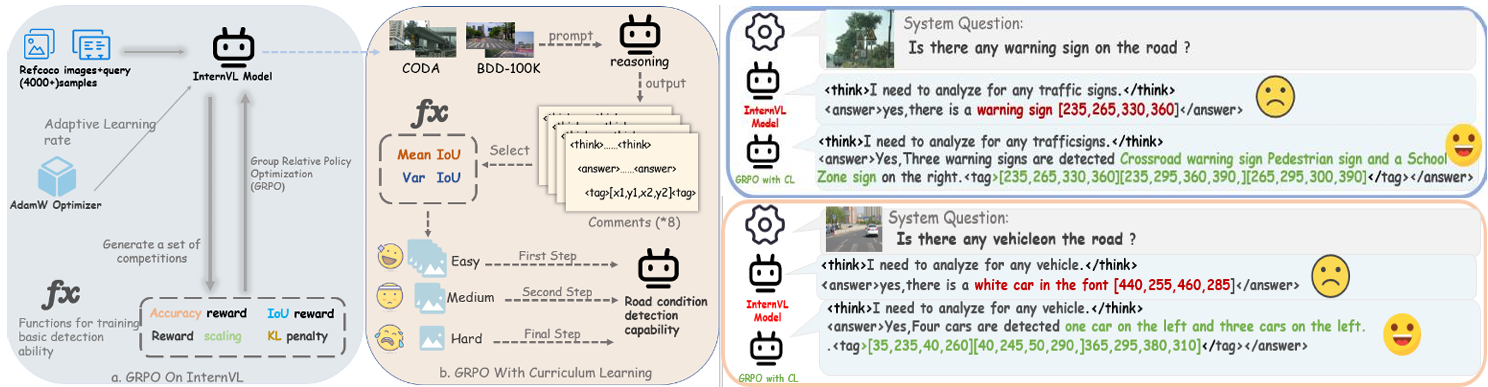}
\caption*{Figure 2-1: Our training pipeline begins with RefCOCO image samples, where the InternVL model is optimized using GRPO with Accuracy-IoU as the reward, establishing fundamental grounding capabilities. Subsequently, CODA and BDD-100K datasets are introduced. For each sample, the model generates eight candidate outputs, which are filtered via mean and variance of IoU to assess difficulty levels. A curriculum learning scheme is then applied, progressively advancing from easier to harder subsets. Finally, the trained model is evaluated on real-world test sets.}
\label{fig:architecture}
\end{figure*}
noisy, providing weak supervision for effective learning. Moreover, the absence of principled data difficulty scheduling exposes models to overly complex samples early in training, leading to suboptimal convergence. Curriculum 
learning offers a compelling alternative[8]—by gradually introducing samples in order of increasing difficulty, it enables models to build capabilities incrementally, thereby improving learning efficiency and robustness.\\\\
In this work, we present the following key contributions:\\\\
1.	We introduce a reinforcement learning framework that augments GRPO with a principled curriculum learning mechanism, enabling more stable and efficient training of MLLMs for detection-centric multimodal tasks.\\\\
2.	We develop a difficulty-aware and semantics-guided data filtering pipeline that dynamically prioritizes high-quality, informative samples during training, reducing susceptibility to annotation noise and distributional shifts.\\\\
3.	We validate our approach on realistic autonomous driving scenarios, achieving consistent \\\\
improvements in localization accuracy and model reliability compared to strong baselines: specifically, a 9.4 \% gain in IoU for pedestrian and non-motorized vehicle detection on the BDD-100K validation set[9], and a 7.1\% increase in IoU for  traffic participants detection on CODA[10].


%% file: 02_related.tex
\section{Related Work}
\label{sec:related}

\subsection{Multimodal Large Language Models}
MLLMs represent a significant advancement in artificial intelligence, extending the capabilities of traditional language models to jointly process and understand multiple modalities. By integrating visual, textual, and sometimes auditory inputs, MLLMs enable more holistic scene understanding. Models such as the LLaVA series[11] combine a pre-trained vision encoder with a LLM, enabling vision-based instruction following—where natural language commands guide image analysis and reasoning. InternVL[12] marks another milestone, achieving strong performance across vision-language benchmarks through large-scale multimodal pretraining and architectural refinement. The release of GPT-4V[13] further demonstrates the potential of MLLMs in complex visual reasoning tasks.\\
In object detection, All-seeing[14] enhances MLLM representation by incorporating structured data, narrowing the performance gap with conventional detectors. Concurrently, VideoChat-TPO[15] introduces a decoder that maps semantic embeddings from the MLLM to spatial coordinates, enabling direct and interpretable object localization. 
\begin{figure*}[t]
\centering
\includegraphics[width=\textwidth,keepaspectratio]{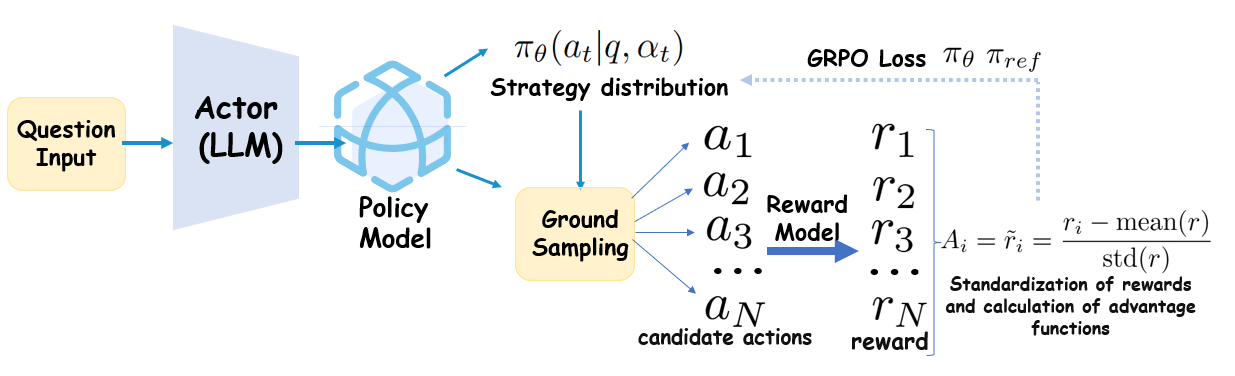}
\caption*{Figure2-2 Overview of GRPO: candidate responses are sampled from the policy distribution, rewards are standardized within each group, and relative advantages guide policy updates under a reference constraint.
}
\label{fig:architecture}
\end{figure*}
With the advent of GRPO, works like Visual-R1[16] leverage reinforcement learning to improve grounding accuracy, significantly advancing MLLMs in cross-modal alignment and precise spatial reasoning.

\subsection{Group Relative Policy Optimization} 
In recent years, GRPO has emerged as a significant advancement in reinforcement learning, particularly 
for addressing the high variance and unstable convergence commonly observed in traditional policy gradient methods when applied to multimodal tasks. Unlike conventional approaches that rely on absolute rewards relative to a global baseline, GRPO introduces an intra-group relative comparison mechanism. Specifically, it constructs a group-wise baseline from multiple candidates generated under the same input, thereby reducing the impact of reward scale variations and improving optimization stability. This relative feedback paradigm enables more robust policy updates in environments with sparse or noisy rewards, while preserving diverse exploration. As a result, GRPO enhances the model’s ability to capture complex cross-modal semantics and spatial correspondences. Owing to these advantages, GRPO has been successfully applied to optimize multimodal large language models, with notable successes such as Visual-R1 demonstrating superior performance over standard supervised fine-tuning (SFT) and vanilla PPO[17] in grounding and visual reasoning tasks.

\subsection{Object Detection for Autonomous Driving}
Object detection in autonomous driving scenarios faces unique and significant challenges that go beyond standard detection benchmarks[18,19]. The dynamic nature of real-world environments, coupled with the wide variety of traffic participants, leads to a pronounced long-tail distribution of object categories, where rare classes are severely underrepresented. Furthermore, practical deployment is hindered by frequent occlusions, cluttered backgrounds, and adverse weather conditions (e.g., fog, rain, or low illumination), all of which degrade the robustness and generalization of conventional detection models. At the same time, strict latency constraints demand efficient, real-time inference under limited computational budgets, posing a critical challenge for model design and optimization.\\
In multimodal detection settings—where models leverage both visual and linguistic inputs—existing approaches often rely on single, hand-crafted supervision signals, limiting their ability to exploit language priors for fine-grained spatial localization. Moreover, training such models is typically unstable due to sparse and noisy reward signals, especially when integrating reinforcement learning paradigms. These issues collectively contribute to high rates of false positives and missed detections, and impede effective transfer across diverse driving environments.\\
To address these limitations, this work proposes a novel optimization framework that integrates GRPO with curriculum-based data selection[8,20]. By leveraging relative feedback within candidate groups, our approach mitigates the instability caused by reward sparsity, while the curriculum strategy progressively exposes the model to increasingly challenging samples. This enables improved robustness and accuracy, particularly for tail classes and complex perceptual conditions, offering a scalable solution for perception systems in autonomous driving.

%% file: 03_method.tex
\section{Method}
\label{sec:method}
{\large\textbf{3.1 GRPO on InternVL}}
\\In InternVL, object detection is formulated as a sequence generation task, where the model autoregressively generates a token sequence encoding bounding box coordinates. To construct a candidate set, the model performs n stochastic rollouts using sampling strategies such as top-p or temperature sampling, yielding a diverse set of predicted boxes {y1,…,yn} . Each generated sequence is mapped to a continuous bounding box via:
\[
b = g(y_i) = h_r(y_i)
\]
where $ h_r(y_i)$denotes the final hidden state representation of the i -th generated sequence[21]\\
To further enhance optimization stability, we adopt GRPO which standardizes rewards within each group to obtain relative advantages:
\[
A_i = \frac{\vec{r}_i - \text{mean}(\{r_i\}_{i=1}^G)}{\text{std}(\{r_i\}_{i=1}^G)}

As illustrated in Figure~2-2, the normalized advantages are then incorporated into the GRPO objective, which jointly maximizes expected advantages while constraining the updated policy against a reference distribution:

\[
\max_{\pi_\theta} \mathbb{E}_{o \sim \pi_{o, \text{old}}}(p) \left[ \sum_{i=1}^G \frac{\pi_\theta(o_i)}{\pi_{o, \text{old}}(o_i)} \cdot A_i - \beta D_{\text{KL}}(\pi_\theta \parallel \pi_{\text{ref}}) \right].

We also incorporate a multi-component loss [22] composed of three distinct terms: the IoU loss, the Format loss, and the KL loss. Each component plays a complementary role in guiding the model toward desired behaviors. Specifically, the IoU loss directly optimizes the overlap between predicted and ground-truth regions, promoting accurate spatial localization. The Format loss enforces structural consistency with respect to predefined output formats—such as syntactic or semantic constraints—ensuring that the model’s predictions adhere to the required representation. Finally, the KL divergence loss regularizes the latent space by aligning the predicted distribution with a prior distribution, facilitating better generalization and smoother latent representations.
\[
L_{\text{overall}} = \alpha L_{\text{iou}} + \beta L_{\text{format}} + \gamma L_{\text{KL}}

The overall training objective is therefore a weighted combination of the GRPO objective and these auxiliary losses, carefully balanced to achieve both high precision and structured coherence in the model outputs.

{\large\textbf{3.2 Progressive Data Selection via Curriculum Learning}}\\
In complex autonomous driving scenarios, training samples exhibit significant variations in difficulty and noise distribution. Naively mixing such heterogeneous samples during training often leads to unstable optimization dynamics—particularly in the early stages—manifesting as convergence oscillations and sparse reward signals.\\
To mitigate these challenges, we propose a curriculum learning-based data screening strategy that progressively exposes the model to samples in an order determined by their estimated difficulty. Our approach leverages a base model pre-trained on large-scale referring expression comprehension benchmarks (RefCOCOand RefCOCO+)[23,24] to assess the difficulty of samples in the target autonomous driving datasets (BDD-100K and CODA).\\

Specifically, for each input instance, we generate multiple detection candidates using stochastic inference, and compute a consistency score based on the spatial overlap and semantic coherence across these predictions. Formally, given a prediction set $T_{pred}$  and the corresponding ground truth T , the overlap is measured by the intersection-over-union:
\[
\text{IoU}(T_{\text{pred}}, T) = \frac{|T_{\text{pred}} \cap T|}{|T_{\text{pred}} \cup T|}
\]
Beyond spatial alignment, we introduce a difficulty estimator D, hat integrates prediction variance V with semantic plausibility $P_{sem}$ :

\[
S_{\text{diff}} = D(P_{\text{sem}}, V)
\]
where higher variance and lower semantic confidence indicate greater ambiguity. Samples $S_{diff}$ are sorted by    and partitioned into progressive curriculum stages.\\
uring training, stage transitions are controlled by a decaying schedule that gradually reduces the weight of easy samples and shifts focus toward harder ones. Let m(t) denote the contribution of easy samples at iteration t:
\[
m(t) = \begin{cases} 
m_0\left(1 - \frac{t}{wT}\right), & 0 \leq t \leq wT, \\ 
0, & t > wT, 
\end{cases}
\]
where T is the total number of iterations and w∈(0,1) adjusts the warm-up duration. This mechanism ensures that early optimization receives stable signals from low-variance samples, while later stages emphasize robustness under challenging, high-variance conditions.\\
 This progressive learning paradigm not only stabilizes training dynamics but also alleviates the problem of reward sparsity by ensuring that the model receives meaningful feedback early in optimization. Importantly, our curriculum design is fully automatic and does not require manual annotation of difficulty levels, making it scalable and adaptable to diverse real-world driving conditions.\\

%% file: 04_experi.tex
\section{EXPERIMENT}
\label{sec:related}

{\large\textbf{4.1 Experiment Settings}}\\
We train the model using the AdamW optimizer[25] with an initial learning rate of 5e−7 and a batch size of 32, for a total of 5,000 training steps. The learning rate is warmed up linearly over the first 10\% of the training schedule and then decayed using a cosine annealing strategy[27].

{\large\textbf{4.2 Data Expansion}}

\textbf{4.2.1 Improved RefCOCO Detection Performance} To further validate the effectiveness of GRPO, we compare it against a baseline that relies solely on Supervised Fine-Tuning (SFT), without reinforcement learning. Experimental results on the standard RefCOCO dataset—across the val, testA, and testB splits—demonstrate consistent and significant improvements. Specifically, the GRPO-optimized model achieves IoU@0.5 gains of 2.1\%, 3.1\%, and 2.5\%, respectively, over the SFT-only counterpart. These improvements go beyond mere memorization of training labels, indicating enhanced generalization across diverse referring expressions and visual contexts.(table~4-1)\\
Qualitative analysis reveals that GRPO encourages the model to better align fine-grained linguistic elements with their corresponding visual regions, leading to more accurate and semantically coherent predictions. This suggests that reinforcement optimization does not merely amplify existing biases in the supervision signal, but instead promotes a deeper understanding of cross-modal semantics—mapping nuanced language descriptions to precise spatial layouts.\\
These results highlight the potential of reinforcement learning as a powerful paradigm for refining multimodal grounding models, particularly in scenarios where pixel-level supervision is sparse or ambiguous, and where robust generalization is critical. The performance gains underscore that GRPO enhances not only detection accuracy but also the model’s ability to reason compositionally across vision and language modalities.\\
\begin{table*}[ht]
\centering
\captionsetup{font=small,labelfont=bf}
\caption{Comparison of IoU@0.5 on referring expression benchmarks. 
Our Visual-R1 achieves consistent improvements across datasets.}
\label{tab:ref_performance}

\renewcommand{\arraystretch}{1.25}
\setlength{\tabcolsep}{16pt} 

\begin{tabular}{lcccccc}
\toprule
\multirow{2}{*}{Model} & 
\multicolumn{2}{c}{RefCOCO IoU@0.5} & 
\multicolumn{2}{c}{RefCOCOg IoU@0.5} & 
\multicolumn{2}{c}{RefCOCO+ IoU@0.5} \\
\cmidrule(lr){2-3} \cmidrule(lr){4-5} \cmidrule(lr){6-7}
& val & test & val & test & val & test \\
\midrule
OFA            & 77.8 & 78.1 & 73.5 & 73.2 & 72.4 & 72.1 \\
BLIP-2         & 79.6 & 80.1 & 74.8 & 75.2 & 74.1 & 74.5 \\
LLaVA         & 80.3 & 80.9 & 75.5 & 75.9 & 74.8 & 75.2 \\
VideoChat-TPO & 81.2 & 81.5 & 76.2 & 76.7 & 75.9 & 76.1 \\
SFT                     & 82.0 & 82.4 & 77.1 & 76.6 & 77.4 & 76.8 \\
\rowcolor{gray!10}
\textbf{GRPO with CL}         & \textbf{85.1} & \textbf{85.5} & \textbf{80.2} & \textbf{80.6} & \textbf{79.3} & \textbf{79.8} \\
\bottomrule
\end{tabular}

\vspace{2mm}
\footnotesize{tabel 4-1: IoU@0.5 is reported on validation and test splits of RefCOCO, RefCOCOg, and RefCOCO+
our GRPO with CL consistently outperforms existing baselines across all benchmarks.}
\end{table*}

\textbf{4.2.2 Curriculum Learning-Based Data Expansion} \\
After establishing foundational vision–language grounding abilities on general referring expression benchmarks (e.g., RefCOCO), we extend training to autonomous driving scenarios to evaluate and enhance real-world applicability. To this end, we incorporate two specialized datasets: CODA, a large-scale benchmark for multi-category detection of traffic participants under diverse weather and lighting conditions, and BDD-100K, a widely used autonomous driving dataset with rich scene variability. These datasets collectively cover critical perception tasks in self-driving systems, including pedestrian, vehicle, and cyclist localization, as well as robust recognition under challenging environments such as nighttime and adverse weather.\\
In contrast to conventional approaches [23] that mix domain-specific data uniformly throughout training, we adopt a curriculum-style data scheduling strategy tailored for GRPO optimization. Directly exposing the model to highly ambiguous or cluttered scenes in early training often destabilizes policy updates due to sparse or inconsistent reward signals. To mitigate this, our curriculum begins with samples exhibiting clear semantics and well-defined object boundaries [28]—approximately 40\% of the training data—ensuring stable initial learning and enabling the policy to establish robust vision–language alignment in spatial reasoning.\\
As training progresses, increasingly complex samples are introduced, including those with heavy occlusions, motion blur, long-tail distributions, and low-visibility conditions. The data composition is dynamically adjusted at every K training steps, smoothly shifting the focus toward CODA samples (from 60\% to 80\%) while maintaining sufficient exposure to BDD-100K for generalization across varied urban environments. This scheduling reflects the higher frequency and variability of traffic participants in real driving contexts, while still preserving diverse scene coverage for robust performance.\\
Table4-2 summarizes the performance trajectory under this curriculum design. Compared to the SFT baseline, GRPO with curriculum learning yields consistent improvements across training steps, reaching 58.9 IoU on CODA and 78.4 IoU on BDD-100K at 5K steps. These gains indicate not only enhanced robustness under complex driving conditions but also faster and more stable convergence. In particular, the improvements are more pronounced at later stages, suggesting that progressive exposure to difficult samples enables the model to effectively internalize long-tail and noisy patterns that conventional uniform sampling fails to capture.\\
\begin{table}[ht]
\centering

\label{tab:coda_cctsdb_curriculum}
\resizebox{\linewidth}{!}{
\begin{tabular}{l|ccc|ccc}
\toprule
\multirow{2}{*}{Method} & \multicolumn{3}{c|}{\textbf{CODA (IoU ↑)}} 
                        & \multicolumn{3}{c}{\textbf{CCTSDB (IoU ↑)}} \\
\cmidrule(lr){2-4} \cmidrule(lr){5-7}
 & Early Stage & Mid Stage & Final Stage & Early Stage & Mid Stage & Final Stage \\
\midrule
SFT & 50.1 & 53.7 & 57.2 & 62.5 & 66.1 & 71.3 \\
GRPO (Curriculum Learning) & 54.8 & 60.2 & \textbf{66.6} & 66.8 & 72.5 & \textbf{78.4} \\
\bottomrule
\end{tabular}}
\\[2pt]
\footnotesize{table 4-2 Comparison of SFT vs. GRPO on autonomous driving datasets using a curriculum learning strategy.}
\end{table}
Our experiments demonstrate that the proposed curriculum-based data expansion significantly enhances the model’s robustness in cross-scenario detection tasks. On the CODA validation set, the model achieves a 7.1\% absolute improvement in mAP for critical traffic participants such as pedestrians and non-motorized vehicles—categories that are often occluded or appear under low-visibility conditions. Meanwhile, on CCTSDB, the model shows a 9.4\% increase in mean IoU for detection in special weather, which are particularly challenging due to their limited pixel footprint and high inter-class similarity.\\
More importantly, compared to a baseline that mixes CODA and BDD-100K data uniformly from the start, our curriculum strategy effectively suppresses gradient instability during training. As shown in the optimization trajectory, the policy gradients exhibit significantly reduced variance, leading to smoother convergence and a more stable learning process [29]. The reinforcement reward curve stabilizes after approximately 5K training steps, indicating that the model has effectively internalized the alignment between natural language descriptions and spatially grounded visual concepts.\\
This marked reduction in training oscillations underscores the effectiveness of gradually increasing task complexity in reinforcement-driven domain adaptation. By exposing the model to clean, unambiguous samples early and progressively introducing noisy, ambiguous, and long-tail instances, the curriculum design mitigates reward sparsity and provides consistent learning signals throughout optimization.\\

These results validate that structured data scheduling is not only beneficial for performance gains but also crucial for training stability in multimodal reinforcement learning. Together, they establish a solid foundation for transferring vision-language models to real-world autonomous driving environments—where robustness, generalization, and reliable convergence are paramount.

{\large\textbf{4.3 SAblation Studys}}

\textbf{4.3.1 Albation Study on Curriculum Learning} \\
To systematically evaluate the contribution of curriculum learning in multimodal object detection and cross-domain generalization, we conduct a comprehensive ablation study with carefully designed training variants. We compare a progressive curriculum learning (CL) strategy—where samples are introduced in increasing order of difficulty from Easy to Hard—against several strong baselines: a random mixing approach that uniformly shuffles all samples regardless of difficulty, extreme regimes that train exclusively on either Easy or Hard subsets, and a full-set direct training scheme that uses all data from the beginning without any prioritization. To ensure an objective and reproducible difficulty assignment, we construct a composite difficulty scoring function [30] based on object size, occlusion level, and linguistic description complexity—three key factors that influence detection challenge in real-world autonomous driving scenarios. Using this metric, we stratify the CODA and BDD-100K datasets into three well-defined difficulty tiers: Easy, Medium, and Hard, enabling a structured, stage-wise training progression. All models are trained under the same optimization protocol and evaluated across multiple dimensions, including detection accuracy (mAP, mIoU), convergence speed, small-object recognition, and cross-scenario robustness. Experimental results demonstrate that progressive CL significantly outperforms all baselines. On the CODA validation set, it achieves a 8.4\% absolute improvement in mAP for pedestrians and non-motorized vehicles compared to random mixing, while it improves mIoU by 5.3\% for small-scale traffic signs—categories that are particularly susceptible to localization errors. Crucially, the CL strategy accelerates early-stage convergence by providing clean, high-signal examples first, leading to more stable gradient updates and faster alignment between vision and language modalities. At the same time, it mitigates overfitting and reward instability in later stages by deferring exposure to highly ambiguous or noisy samples until the model has acquired sufficient representational capacity. In contrast, models trained solely on Easy data fail to generalize to complex real-world conditions, while those exposed only to Hard samples exhibit erratic optimization behavior and poor final performance, confirming that unbalanced data exposure harms both stability and generalization. The progressive, difficulty-aware scheduling inherent in CL enables the model to build a hierarchical knowledge structure—starting from basic visual grounding and gradually advancing toward robust reasoning under occlusion, clutter, and linguistic ambiguity. This staged learning process proves especially effective in handling long-tail class distributions and domain shifts, highlighting the importance of structured data ordering in multimodal reinforcement learning. Together, these findings validate the theoretical and practical value of curriculum learning in optimizing vision-language models for real-world autonomous driving, where robustness, efficiency, and generalization are paramount. \\

\begin{figure}[H]
\centering
\includegraphics[width=\linewidth,keepaspectratio]{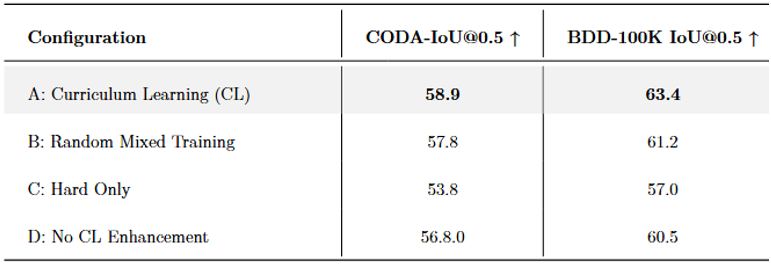}
\captionsetup{font=scriptsize} 
\caption*{table 4-3 Curriculum Learning yields consistent improvements across datasets, while other strategies either underfit or generalize poorly.}
\label{fig:example}
\end{figure}

\textbf{4.3.2 Ablation Study on Training Hyperparameters} \\
To further investigate the impact of training hyperparameters on optimization dynamics and cross-scenario detection performance, we conduct a series of four core ablation studies, systematically analyzing the effects of learning rate sensitivity, IoU reward weighting, KL regularization strength, and curriculum phase duration.
First, in the learning rate sensitivity analysis, we perform a grid search over different learning rate configurations. The results reveal that the learning rate plays a pivotal role in balancing convergence stability and final accuracy. An excessively high learning rate leads to unstable training [31], with pronounced oscillations in the loss curve and a significant increase in formatting errors due to overly aggressive parameter updates. Conversely, an overly conservative learning rate results in slow convergence—although it may yield slightly more stable generalization in certain scenarios, the overall performance is suboptimal due to insufficient optimization progress within the fixed 5K-step schedule. Our experiments demonstrate that a moderate learning rate (5e-7)achieves the best trade-off, enabling rapid yet stable convergence and allowing the model to reach strong performance within limited training steps.\\

\begin{figure}[H]
\centering
\includegraphics[width=\linewidth,keepaspectratio]{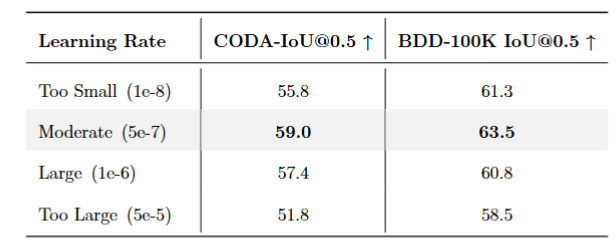}
\captionsetup{font=scriptsize} 
\caption*{table 4-4 Excessively small learning rate slows convergence, while overly large values destabilize training.}
\label{fig:example}
\end{figure}

Second, we examine the influence of IoU reward weighting on detection accuracy, particularly for small objects and complex scenes. By varying the coefficient of the IoU term in the reward function, we observe a clear trend: low IoU weights provide insufficient geometric supervision [24], causing the model to drift in spatial localization and perform poorly on small targets. As the weight increases, both small-object IoU and cross-dataset robustness improve significantly, indicating stronger alignment between predicted and ground-truth regions. However, beyond a certain threshold, excessive emphasis on IoU leads to reward dominance that distorts the policy update, resulting in unstable decoding behavior and increased format violations—such as malformed bounding box coordinates or invalid syntactic structures. This non-linear relationship underscores the importance of carefully calibrating the IoU reward to enhance detection precision without compromising output coherence.\\
\begin{figure}[H]
\centering
\includegraphics[width=\linewidth,keepaspectratio]{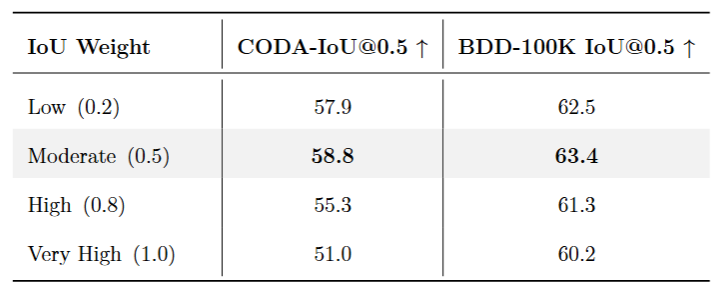}
\captionsetup{font=scriptsize} 
\caption*{table 4-5 Too small IoU weight weakens geometric constraints; too large emphasizes precision but harms output stability. Moderate weight achieves balance.}
\label{fig:example}
\end{figure}

Third, we perform a systematic ablation on the KL penalty strength to assess its role in stabilizing policy updates and regulating exploration. A weak KL coefficient allows the policy to deviate too far from the reference model during optimization [20], leading to large reward variance, frequent training instability, and in some cases, catastrophic performance drops. On the other hand, an overly strong KL penalty overly constrains the policy update, limiting the model’s ability to explore beneficial directions and ultimately resulting in underfitting. Our findings show that a moderate KL regularization strength strikes the optimal balance between exploration and exploitation: it effectively suppresses reward variance, stabilizes the training trajectory, and promotes consistent improvements across domains, particularly in out-of-distribution scenarios where reliable generalization is critical.\\

\begin{figure}[H]
\centering
\includegraphics[width=\linewidth,keepaspectratio]{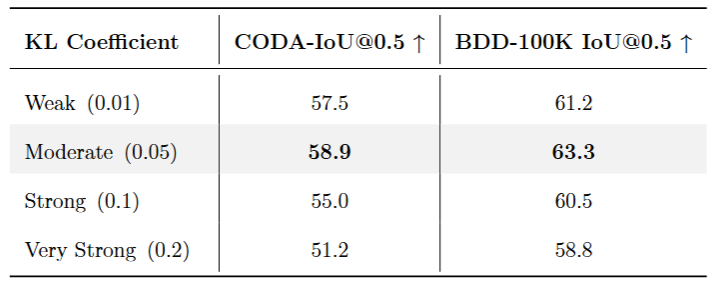}
\captionsetup{font=scriptsize} 
\caption*{table 4-6  Weak KL leads to unstable policy updates, strong KL suppresses exploration. Moderate KL finds the optimal trade-off..}
\label{fig:example}
\end{figure}
Finally, we analyze the impact of curriculum phase length—the duration spent at each difficulty stage—on the model’s learning dynamics. In our curriculum framework, data is introduced from easy to hard, and the length of each phase governs the pace of knowledge acquisition. Short phases(0.5K steps) acelerate exposure to hard samples but risk destabilizing training [32] by introducing high-noise, ambiguous instances too early, thereby impairing overall generalization. In contrast, excessively long phases(2-2.5K steps) allow thorough learning on easy data but delay adaptation to complex scenes, leading to slower progression and reduced readiness for real-world challenges. Experimental results indicate that intermediate phase(1-1.5K) lengths—sufficiently long to consolidate foundational skills yet flexible enough to transition in a timely manner—facilitate the most effective knowledge transfer. This balanced pacing enables the model to gradually build robust detection capabilities, achieving superior performance on small objects, occluded instances, and linguistically complex queries.\\
\begin{figure}[H]
\centering
\includegraphics[width=\linewidth,keepaspectratio]{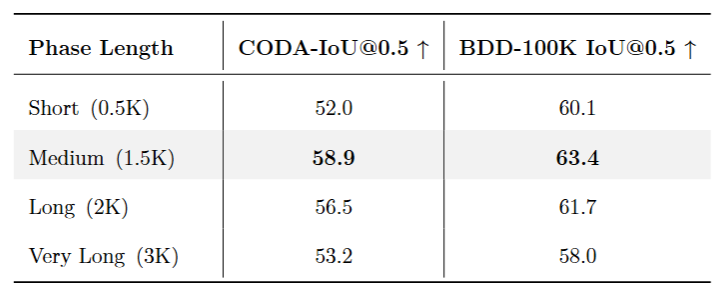}
\captionsetup{font=scriptsize} 
\caption*{table 4-7   Short stages expose the model too early to noise; very long stages slow adaptation. Moderate stage length provides the best balance.}
\label{fig:example}
\end{figure}

In summary, these four ablation studies collectively validate the critical role of training hyperparameters in shaping optimization stability, convergence speed, and cross-scenario generalization. Learning rate, reward 
weighting, regularization strength, and curriculum scheduling do not act in isolation; rather, they interact synergistically to determine the overall effectiveness of the training pipeline. Properly tuned, they can significantly enhance both detection accuracy and training robustness; when misconfigured, they risk performance degradation and unstable learning. These findings provide clear practical guidelines for large-scale vision-language model optimization and offer theoretical insights into the design of effective training strategies for multimodal detection in real-world autonomous systems.\\

%% file: 10_conclusion.tex
\section{Conclusion}
\label{sec:conclusion}

This paper presents a novel cross-modal object detection framework based on Multimodal Large Language Models (MLLMs) and Guided Reinforcement Policy Optimization (GRPO), with systematic validation of its effectiveness in autonomous driving scenarios. By first pre-training on RefCOCO and RefCOCO+, the model acquires robust language-guided localization capabilities, establishing a strong foundation for vision-language grounding. Subsequently, through a curriculum-based data expansion strategy on domain-specific datasets—CODA for multi-category traffic participant detection and CCTSDB for fine-grained traffic sign recognition—the model progressively adapts to complex urban environments and long-tail class distributions. This staged learning approach mitigates the challenge of reward sparsity in reinforcement learning, stabilizes training dynamics, and enhances generalization under real-world conditions.\\
To further improve optimization and prediction quality, we design a composite reward function that jointly optimizes IoU accuracy and output format correctness, augmented with an intra-group comparison mechanism that amplifies meaningful reward signals. This enables GRPO to guide the policy update more effectively, leading to faster convergence and greater interpretability in bounding box generation. Unlike traditional detectors [33]or purely supervised MLLM baselines [34], our method explicitly learns to align linguistic semantics with spatial geometry through reinforcement, rather than relying solely on static supervision.\\
Extensive experiments demonstrate that our approach consistently outperforms existing methods across multiple datasets, object categories, and environmental conditions. Particularly notable gains are observed in detecting small objects, handling cluttered backgrounds, and recognizing rare or ambiguous classes—scenarios that are notoriously challenging for conventional models [35]. The results confirm that reinforcement-driven optimization does not merely fine-tune existing capabilities but enables a deeper, more adaptive integration of vision and language modalities.
Overall, this work demonstrates that reinforcement learning can serve as a powerful engine for transferring multimodal models from basic comprehension to complex, real-world perception tasks. By combining curriculum design, reward shaping, and policy regularization, we establish a scalable and effective paradigm for vision-language grounding in autonomous driving[36]. Our findings open promising directions for building more robust, interpretable, and generalizable perception systems through interactive, feedback-driven learning.\\

%% file: 12_appendix.tex
\section{Appendix Section}
\label{sec:appendix_section}
Supplementary material goes here.

%% file: _main.bbl
\begin{thebibliography}{10}

\bibitem{1}
A.~Radford, J.~W.~Kim, et al.
\newblock Learning transferable visual models from natural language supervision.
\newblock In {\em ICML}, 2021.

\bibitem{2}
J.-B.~Alayrac, J.~Donahue, et al.
\newblock Flamingo: a visual language model for few-shot learning.
\newblock In {\em NeurIPS}, 2022.

\bibitem{3}
H.~Bao, W.~Wang, et al.
\newblock VLMo: Unified vision-language pre-training with mixture-of-modality-experts.
\newblock In {\em NeurIPS}, 2022.

\bibitem{4}
W.~Wang, H.~Bao, et al.
\newblock Image as a foreign language: BEiT pretraining for all vision and vision-language tasks.
\newblock In {\em CVPR}, 2023.

\bibitem{5}
R.~Hu, A.~Singh, et al.
\newblock UniT: Multimodal multitask learning with a unified transformer.
\newblock In {\em ICCV}, 2021.

\bibitem{6}
Y.~Zhang, K.~Han, et al.
\newblock Meta-Transformer: A unified framework for multimodal learning.
\newblock {\em arXiv:2307.10802}, 2023.

\bibitem{7}
A.~Dosovitskiy, L.~Beyer, A.~Kolesnikov, D.~Weissenborn, X.~Zhai, T.~Unterthiner, M.~Dehghani, M.~Minderer, G.~Heigold, S.~Gelly, J.~Uszkoreit, and N.~Houlsby.
\newblock An image is worth 16x16 words: Transformers for image recognition at scale.
\newblock In {\em ICLR}, 2021.

\bibitem{8}
Y.~Bengio, J.~Louradour, et al.
\newblock Curriculum learning.
\newblock In {\em ICML}, 2009.

\bibitem{9}
F.~Yu, H.~Chen, X.~Wang, W.~Xian, Y.~Chen, F.~Liu, V.~Madhavan, and T.~Darrell.
\newblock BDD100K: A diverse driving dataset for heterogeneous multitask learning.
\newblock In {\em CVPR}, pages 2636--2645, 2020.

\bibitem{10}
K.~Li, K.~Chen, H.~Wang, L.~Hong, C.~Ye, J.~Han, Y.~Chen, W.~Zhang, C.~Xu, D.-Y.~Yeung, X.~Liang, Z.~Li, and C.~Xu.
\newblock CODA: A real-world road corner case dataset for object detection in autonomous driving.
\newblock {\em arXiv:2203.07724}, 2022.

\bibitem{11}
H.~Liu, C.~Li, et al.
\newblock Visual instruction tuning.
\newblock In {\em NeurIPS}, 2023.

\bibitem{12}
Z.~Chen, J.~Wu, et al.
\newblock InternVL: Scaling up vision foundation models and aligning for generic visual-linguistic tasks.
\newblock {\em arXiv:2312.14238}, 2023.

\bibitem{13}
OpenAI.
\newblock GPT-4V(ision) system card.
\newblock 2023.

\bibitem{14}
W.~Wang et al.
\newblock All-Seeing: A foundation model for object perception.
\newblock {\em arXiv:2308.09257}, 2023.

\bibitem{15}
H.~Shen, P.~Liu, J.~Li, C.~Fang, Y.~Ma, J.~Liao, Q.~Shen, Z.~Zhang, K.~Zhao, Q.~Zhang, R.~Xu, and T.~Zhao.
\newblock VLM-R1: A stable and generalizable R1-style large vision-language model.
\newblock {\em arXiv:2504.07615}, 2025.

\bibitem{16}
A.~Dao and D.-B.~Vu.
\newblock AlphaMaze: Enhancing large language models' spatial intelligence via GRPO.
\newblock {\em arXiv:2502.14669}, 2025.

\bibitem{17}
Z.~Chen, R.~Niu, H.~Kong, and Q.~Wang.
\newblock TGRPO: Fine-tuning vision-language-action model via trajectory-wise group relative policy optimization.
\newblock {\em arXiv:2506.08440}, 2025.

\bibitem{18}
H.~Tan and J.~Pan.
\newblock GTPO and GRPO-S: Token and sequence-level reward shaping with policy entropy.
\newblock {\em arXiv:2508.04349}, 2025.

\bibitem{19}
Y.~Chao, J.~Liu, J.~Tang, and G.~Wu.
\newblock AnomalyR1: A GRPO-based end-to-end MLLM for industrial anomaly detection.
\newblock {\em arXiv:2504.11914}, 2025.

\bibitem{20}
J.~Schulman, F.~Wolski, et al.
\newblock Proximal policy optimization algorithms.
\newblock {\em arXiv:1707.06347}, 2017.

\bibitem{21}
H.~Caesar, V.~Bankiti, et al.
\newblock nuScenes: A multimodal dataset for autonomous driving.
\newblock In {\em CVPR}, 2020.

\bibitem{22}
P.~Sun, H.~Kretzschmar, et al.
\newblock Scalability in perception for autonomous driving: Waymo open dataset.
\newblock In {\em CVPR}, 2020.

\bibitem{23}
P.~Soviany, R.~T.~Ionescu, et al.
\newblock Curriculum learning: A survey.
\newblock {\em IJCV}, 2022.

\bibitem{24}
H.~Rezatofighi, N.~Tsoi, et al.
\newblock Generalized intersection over union: A metric and a loss for bounding box regression.
\newblock In {\em CVPR}, 2019.

\bibitem{25}
L.~Yu, P.~Poirson, et al.
\newblock Modeling context in referring expressions.
\newblock In {\em ECCV}, 2016.

\bibitem{26}
J.~Mao, J.~Huang, et al.
\newblock Generation and comprehension of unambiguous object descriptions.
\newblock In {\em CVPR}, 2016.

\bibitem{27}
I.~Loshchilov and F.~Hutter.
\newblock SGDR: Stochastic gradient descent with warm restarts.
\newblock In {\em ICLR}, 2017.

\bibitem{28}
L.~Jiang, D.~Meng, et al.
\newblock Difficulty-aware attention network with curriculum learning for medical image segmentation.
\newblock In {\em MICCAI}, 2018.

\bibitem{29}
D.~Weinshall, G.~Cohen, and D.~Amir.
\newblock Curriculum learning by transfer learning: Theory and experiments with deep networks.
\newblock In {\em ICML}, 2018.

\bibitem{30}
G.~Hacohen and D.~Weinshall.
\newblock On the power of curriculum learning in training deep networks.
\newblock In {\em ICML}, 2019.

\bibitem{31}
I.~Goodfellow, Y.~Bengio, and A.~Courville.
\newblock {\em Deep Learning}.
\newblock MIT Press, 2016.

\bibitem{32}
A.~Graves, M.~G.~Bellemare, et al.
\newblock Automated curriculum learning for neural networks.
\newblock In {\em ICML}, 2017.

\bibitem{33}
S.~Ren, K.~He, et al.
\newblock Faster R-CNN: Towards real-time object detection with region proposal networks.
\newblock In {\em NeurIPS}, 2015.

\bibitem{34}
J.~Li, D.~Li, et al.
\newblock BLIP-2: Bootstrapping language-image pre-training with frozen image encoders and large language models.
\newblock In {\em ICML}, 2023.

\bibitem{35}
M.~Kisantal, Z.~Wojna, et al.
\newblock Augmentation for small object detection.
\newblock {\em arXiv:1902.07296}, 2019.

\bibitem{36}
W.~Zeng, M.~Liang, et al.
\newblock A survey of vision-language pre-training for autonomous driving.
\newblock {\em IEEE Transactions on Intelligent Transportation Systems}, 2023.

\end{thebibliography}
